\documentclass[letterpaper, 10 pt, conference]{ieeeconf}  % Comment this line out if you need a4paper

\IEEEoverridecommandlockouts                              % This command is only needed if 
\usepackage{amsmath}
\usepackage{amssymb}
\usepackage{graphicx}
\usepackage{amsmath} % assumes amsmath package installed
\usepackage{amssymb}  % assumes amsmath package installed
\usepackage{graphicx}
\usepackage{bbm}
\usepackage{xcolor}
\usepackage{hyperref}
\usepackage{todonotes}
\usepackage{subfig}
\usepackage{comment}
\usepackage{multirow}

\usepackage{booktabs}
\usepackage{multirow}
\usepackage{siunitx}
\usepackage{caption}
\usepackage{float}
\usepackage{xcolor}

\usepackage[inline]{trackchanges}
\addeditor{CC} % Chiho

\title{\LARGE \bf
DIDER: Discovering Interpretable Dynamically Evolving Relations}
\author{Enna Sachdeva$^{1}$ and Chiho Choi$^{1}$% <-this % stops a space
\thanks{$^{1}$Honda Research Institute, CA, USA}%
\thanks{\tt\small {esachdeva@honda-ri.com, cchoi@honda-ri.com}}
}

\begin{document}

\maketitle
\thispagestyle{empty}
\pagestyle{empty}

\vspace{-1.8em}

%%%%%%%%%%%%%%%%%%%%%%%%%%%%%%%%%%%%%%%%%%%%%%%%%%%%%%%%%%%%%%%%%%%%%%%%%%%%%%%%
\begin{abstract}
Effective understanding of dynamically evolving multiagent interactions is crucial to capturing the underlying behavior of agents in social systems. It is usually challenging to observe these interactions directly, and therefore modeling the latent interactions is essential for realizing the complex behaviors. Recent work on Dynamic Neural Relational Inference (DNRI) captures explicit inter-agent interactions at every step. However, prediction at every step results in noisy interactions and lacks intrinsic interpretability without post-hoc inspection. Moreover, it requires access to ground truth annotations to analyze the predicted interactions, which are hard to obtain. This paper introduces DIDER, Discovering Interpretable Dynamically Evolving Relations, a generic end-to-end interaction modeling framework with intrinsic interpretability. DIDER discovers an interpretable sequence of inter-agent interactions by disentangling the task of latent interaction prediction into sub-interaction prediction and duration estimation. By imposing the consistency of a sub-interaction type over an extended time duration, the proposed framework achieves intrinsic interpretability without requiring any post-hoc inspection. We evaluate DIDER on both synthetic and real-world datasets. The experimental results demonstrate that modeling disentangled and interpretable dynamic relations improves performance on trajectory forecasting tasks.  
\end{abstract}

%%%%%%%%%%%%%%%%%%%%%%%%%%%%%%%%%%%%%%%%%%%%%%%%%%%%%%%%%%%%%%%%%%%%%%%%%%%%%%%%

\begin{figure*}[hbt!]
    \centering
    \subfloat[\centering Interactions Prediction by  DNRI]{{\includegraphics[width=0.30\linewidth]{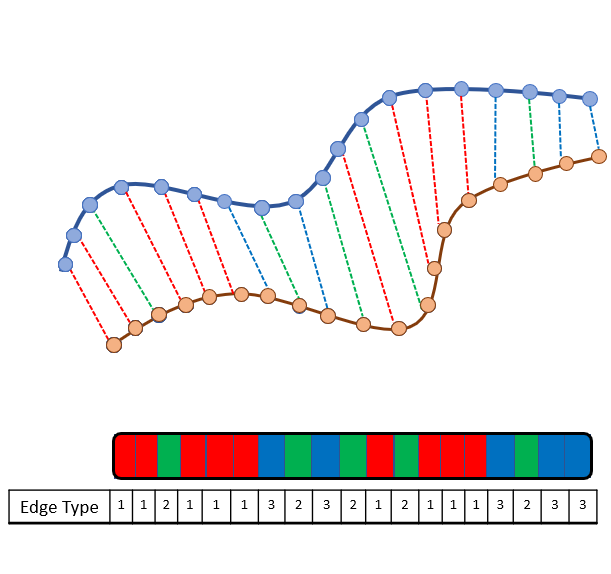} }\label{fig:toy_joint_vanilla_vae}}
    \qquad
    \subfloat[\centering Interactions prediction by DIDER ]{{\includegraphics[width=0.30\linewidth]{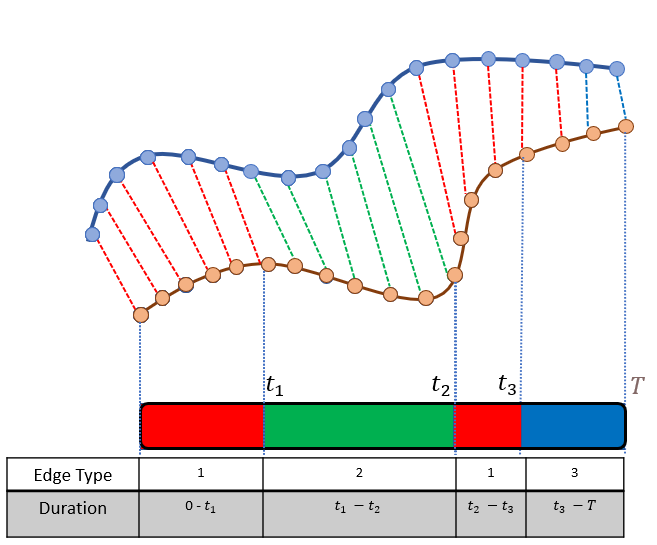} \label{fig:toy_joint_distance_vae}}}
    \caption{Overview of the interaction prediction of DIDER as compared to DNRI. a) DNRI predicts interaction at every time step t, which results in noisy and non-interpretable interactions. b) DIDER disentangles the task of interaction prediction into sub-interaction prediction and duration estimation, which provides interpretable dynamic interactions. By learning duration along with interaction/edge type, DIDER guides the model to learn consistent sub-interactions for extended duration of time.}
    \label{fig:dider_overview}
        \vspace{-1.8em}
\end{figure*}

\section{INTRODUCTION}

Real-world applications such as autonomous driving, mobile robot navigation, and air-traffic management involve multiagent interactions for joint behavior prediction and complex decision making. Modeling these interactions is crucial to understanding the underlying dynamic behavior of the agents. For instance, the future behavior (yielding or right of way) of a vehicle approaching an intersection is influenced by another approaching vehicle. However, it is challenging to model these interagent interactions, as we often do not know about the ground truth interactions between agents.

%Further, its hard to annotate its hard to annotate the ground truth interaction labels at frame level.  

In recent years, there has been a considerable amount of work towards explicit modeling of multiagent interactions from raw trajectories\cite{kipf2018neural, li2020evolvegraph, graber2020dynamic, gong2021memory}. These methods leverage graph neural networks to model relational structures with multiple interaction types. It was firstly introduced in Neural Relational Inference (NRI) \cite{kipf2018neural}, which infers a static relational graph between multiple agents while simultaneously modeling the dynamics of the interacting system. However, most real-world social systems involve dynamically evolving multiagent interactions. To address this gap of modeling dynamic interactions, Dynamic NRI \cite{graber2020dynamic} and Evolvegraph \cite{li2020evolvegraph} were proposed. These models discover unseen interactions between agents at every step to improve the performance of trajectory forecasting tasks. While per-step prediction provides dynamic interactions, it results in a noisy sequence of non-interpretable interactions. These methods usually require post-hoc analysis to interpret these noisy interactions, which could be ambiguous or falsely interpreted by humans. 

%easy PLUG-IN
%One possible reason could be that it is impossible to approach the goal of interpretability without access to the ground truth labels in order to indicate the correspondence between learned representation and human-defined concepts \cite{zhu2021and}.

%it becomes challenging to interpret the inferred relations, without manual inspection the relations, or requiring step-level groundtruth annotations. 

 %in contrast to real-world settings where dynamic interactions can usually be decomposed into a disentangled sequence of sub-interactions. 

In most real-world situations, agents interact with each other in a sequence of sub-interactions for an extended period to jointly execute the downstream task. This is similar to how humans usually tend to break down the long duration of task into a sequence of sub-tasks \cite{sachdeva2021dynamic, sachdeva2021maedys}. For instance, in a lane-changing scenario, a new-follower vehicle (in the new lane) may be required to execute a \textit{yielding} policy to yield to the lane-changing car, followed by a \textit{car-following} policy by maintaining a safe distance from the leader car, to navigate safely. We argue that to discover an interpretable sequence of disentangled sub-interactions, one should account for the extended duration of sub-interaction between agents. This may be achieved by incorporating an additional module to predict the time duration of inter-agent sub-interactions. This additional time duration predictor module adds an interpretability constraint to the model and generate an intrinsically interpretable \cite{li2022online} sequence of sub-interactions. Additionally, these extracted sequences of disentangled sub-interactions facilitate the generation of new and diverse scenarios by combining these interactions in various ways \cite{wang2018extracting},\cite{zhang2020multi}. 

%These methods rely on the model to learn the relation types and their duration, which is a challenging task. 
%Further, achieving per step annotation of ground truth interactions is usually tedious for the expert, biased towards the annotator, and is not scalable. So, the question arises- Can we learn the type of dynamically evolving interactions and their duration from the data, which is more interpretable?  

This paper introduces DIDER- Discovering Interpretable Dynamically Evolving Relations, an unsupervised learning framework for discovering interpretable dynamic multiagent interactions from observations. It leverages VAE-framework to discover interpretable dynamic temporal interactions while simultaneously learning the dynamic model of the system. We incorporate intrinsic interpretability into the model by decoupling the interaction prediction task into sub-interaction and duration predictions. 
The key contributions of this work are summarized as follows:

\begin{itemize}
    \item We propose an end-to-end explicit interaction modeling framework, with intrinsic interpretability, by disentangling dynamic interaction prediction into sub-interaction prediction and duration prediction, as shown in Fig. \ref{fig:dider_overview}. 
    
   \item The proposed model uses trajectory prediction as a surrogate task for learning interpretable dynamically evolving interactions. By predicting each sub-interaction's start and end time, the model provides better interpretability of the latent interactions while improving the performance on downstream trajectory prediction task. 

    \item The proposed model is a generic framework for modeling dynamic interactions and is flexible to be incorporated into any existing VAE-like relational inference framework to improve interactions interpretability and trajectory prediction performance. 
    
    \item We evaluate the performance of the proposed framework on both simulated and realistic trajectory forecasting tasks and visualize the predicted interactions to elucidate the interpretability of the predicted interactions. 
\end{itemize}

We use the terms \textit{relations}, \textit{interactions}, and \textit{edges} interchangeably in this paper.

\section{Related Work}

\subsection{Interpretable Motion Prediction Frameworks}
In recent years, interpretability has been considered an important factor in developing motion prediction frameworks. Recently, Brewitt et al. introduced GRIT \cite{brewitt2021grit}, a goal recognition framework with interpretable decision trees on vehicle trajectory dataset. The encoding of discrete latent space in CVAE framework in motion prediction frameworks like Trajectron \cite{ivanovic2019trajectron} and Trajectron ++ \cite{salzmann2020trajectron++} aids interpretability of the learned latent space. Parth et al. \cite{kothari2021interpretable} combines rule-based models with neural network-based models to predict interpretable high-level intents as well as scene-specific residuals. These methods encode intrinsic interpretability into the model to make black-box models more transparent with or without domain knowledge. However, these methods do not explicitly model the interactions between agents for the motion prediction task. 

\subsection{Interpretable Multiagent Interactions}
Several works have realized interaction modeling and relational reasoning using Graph Neural Networks \cite{kipf2018neural, graber2020dynamic, gong2021memory, li2020evolvegraph, alet2019neural, xiao2020dynamic,  lowe2020amortized}. They introduce nodes to represent the interactive agents and edges to represent their interaction types. By explicitly modeling the dynamic interaction graphs, they learn the dynamic model of the system. While these methods model the agents' underlying static and dynamic interactions, their interpretability was not explored until recently. Recent work on Grounded Relational Inference (GRI) \cite{tang2021grounded} sets as a stepping stone towards generating interpretable and grounded inter-agent interaction graphs. GRI learns the reward functions for various semantically meaningful interactions between agents while simultaneously modeling system dynamics by formulating the problem as Inverse Reinforcement Learning. %It relies on expert domain knowledge to design reward functions corresponding to each interaction type, making it challenging to scale to complex intersection scenarios. 
Another very recent work by Lingfeng and Chen et al. \cite{sun2022domain} leverages pseudo labels to enforce the model to learn interpretable interactions. Further, GRIN \cite{li2021grin} disentangles inter-agent interactions from agents' intentions for better interpretability over inferred \textit{static} interactions. \cite{lee2019joint} partially addresses the interpretability of predicted interactions using a supervised learning framework by generating a simple labeling function to annotate the ground truth interactions between agents. While all these methods primarily focus on improving the interpretability of interactions using domain knowledge, they assume interactions to be static across time. We are the first to address discovering disentangled and interpretable sequences of dynamic interactions from multiagent observations to the best of our knowledge. 

\vspace{-0.2em}

%This was first proposed by Kipf et.al in their work on Neural Relational Inference (NRI)\cite{kipf2018neural}. NRI formulates the problem using a Variational Autoencoder (VAE) framework, without any supervision of underlying relation types. Further, Factorized NRI \cite{webb2019factorised} factorizes the interaction graphs with several edge types into separate layer graphs. However, both these methods assumes the underlying relations to be static across time. Recent work on DNRI \cite{graber2020dynamic} and Evolvegraph \cite{li2020evolvegraph} learns dynamically evolving spatio-temporal relations among agents, by predicting relations at each time step. Further, MemDNRI \cite{gong2021memory} uses memory pools as temporally global latent variables, to capture the long term information in DNRI. While all these methods uses trajectory prediction as surrogate task towards predicting inter-agent relations, none of these methods addresses the interpretability of the inferred latent space, without supervision. In order to interpret the inferred relation types, one needs to get an idea of the frames is a non-trivial task, and may require frame level supervision. This is quite difficult to obtain. One needs to get an idea of the frames, is a non-trivial task, and may require frame level supervision, which is quite difficult to obtain.
\subsection{Trajectory Segmentation}
Trajectory segmentation has been well studied in the literature to facilitate learning localized control policies and combinatorial generalization to unseen scenarios. There exist several unsupervised learning frameworks to decompose trajectories into various skills \cite{shankar2020learning, krishnan2017transition, despinoy2015unsupervised}. These methods aim to decompose trajectories into a sequence of subgoals or skills. TSC-DL \cite{murali2016tsc} segments trajectories into locally-similar contiguous sections but requires prior knowledge of the number of segments. A similar approach CompILE \cite{kipf2019compile}, segments a trajectory in an unsupervised framework but lacks interpretability in latent space and learning length-independent skills. Recent work on SKID \cite{tanneberg2021skid} operates on trajectories with a varying and unknown number of skills per trajectory. Most of the work in this direction addresses single-agent trajectory segmentation to various subgoals or skills. However, none of these methods aim to segment latent multiagent interactions in an unsupervised manner. 

%It also requires a weak supervision of the maximum number of segments. Defining the number of segments in a trajectory is a non-trivial and subjective task, and adds annotator's bias. To address this, a 

%A similar framework COMPILE has been investigated but the latent space is not interpretable. Another method SKID \cite{} has been developed to segment the trajectories into variable number of segments, in an unsupervised way. Our method is losely inspired from SKID. 
%In many real-world tasks, the demonstrations are noisy, there exists a pattern of sequence of segments, which can be exploited to infer high-level understanding of the downstream task.

%%%%%%%%%%%%%%%%%%%%%%%%%%%%%%%%%%%%%%%%%%%%%%%%%%%%%%%%%%%%%%%%%%%%%%%%%%%%%%%%
\section{Background}

\subsection{Dynamic Neural Relational Inference}

Dynamic Neural Relational Inference (DNRI) \cite{graber2020dynamic} models the dynamic evolving relation types between agents by predicting $z^t_{i, j}$ at each time step. It simultaneously learns the dynamic model of the interacting system. DNRI formulates this problem using a Conditional Variational Autoencoder (CVAE) framework. Consider a set of $N$ agents with their trajectories (of duration $T$) denoted as: $x_1^{1:T}, x_2^{1:T},....., x_N^{1:T}$, it predicts the trajectories using relational embeddings. The interactions between entities are represented by $z^t_{i, j} \epsilon \{1, 2, ..., e\}$ for every pair of entities $(i, j)$ at time step $t$, where $e$ denotes the number of possible interaction types between entities.
 
LSTMs are used to model the dynamic prior $p_\phi(z|x)$ and encoder $q_\phi(z|x)$. The encoder at each timestep is conditioned on a full trajectory, while the prior is conditioned on the observation and relation prediction from previous steps: 
% \vspace{-1.5em}
\begin{equation}
\begin{split}
    p_\phi(z|x) :=  \prod_{t=1}^T p_\phi(z^{t}|x^{1:t}, z^{1:t-1}),\\
    q_\phi(z|x) :=  \prod_{t=1}^T q_\phi(z^{t}|x^{1:T}).
\end{split}
\end{equation}
The decoder then predicts the future states of the entities $x$. The decoder is the formulation conditioning on the dynamic $z_t$ sampled from the encoder at every time step $t$.
% \vspace{-1.2em}
\begin{equation}
p_\theta(x|z) :=  \prod_{t=1}^T p_\theta(x^{t+1}|x^{1:t}, z^{1:t}).
\end{equation}
$\theta$, $\phi$ are the trainable parameters of probability distributions, which are optimized by maximizing the following evidence lower bound (ELBO).
% \vspace{-1em}
\begin{equation}
\mathcal{L}(\phi, \theta) = \mathbb{E}_{q_\phi}(z|x)[\log p_\theta(x|z)] - KL[q_\phi(z|x)||p_\phi(z|x)]
\end{equation}

\vspace{-0.8em}
\subsection{Trajectory Segmentation using SKID}
Our method is loosely inspired by SKID \cite{tanneberg2021skid}, an unsupervised framework to segment the trajectories into reoccurring patterns (skills) from unlabelled demonstrations. SKID frames the problem using VAEs, with latent space $z = \{z_d, z_s\}$ describing the properties of a segment, where $z_d$ and $z_s$ represents duration of the skill and skill type. 

SKID models the skill duration $z_d$ with a Gaussian distribution given a prior i.e. $z_d \sim \mathcal{N}(\mu_d, \sigma_d^2)$. Each skill duration $z_d$ is obtained using the remainder of the trajectory, i.e., the part of the trajectory that has not been explained by all previous $z_d$. This extracted sub-trajectory $\tau$ is then used for learning the skill type $z_s$. Assuming that a trajectory consists of $N$ segments, this iterative process is repeated $N$ times until it reaches the last time step of the trajectory. The learning is done by jointly optimizing the generative model and the inference network by maximizing the evidence lower bound. SKID utilizes full trajectories for learning skills and duration, making it suitable for offline settings.

\begin{comment}
as follows- 

\begin{equation}
\begin{split}
\mathcal{L}(\theta, \phi_d, \phi_e) = \mathbb{E}_{q_\phi}(z_e|x)[log p_{\theta_d}(x|z_e)] \\                - \beta_d KL[q_\phi_d(z_d|x)||p(z_d)] \\
- \beta_e KL[q_\phi_e(z_e|x)||p_\phi_e(z_e|x).
\end{split}
\end{equation}
\end{comment}

%%%%%%%%%%%%%%%%%%%%%%%%%%%%%%%%%%%%%%%%%%%%%%%%%%%%%%%%%%%%%%%%%%%%%%%%%%%%%%%%

\begin{figure*}[t]
    \centering
    \includegraphics[width=\linewidth]{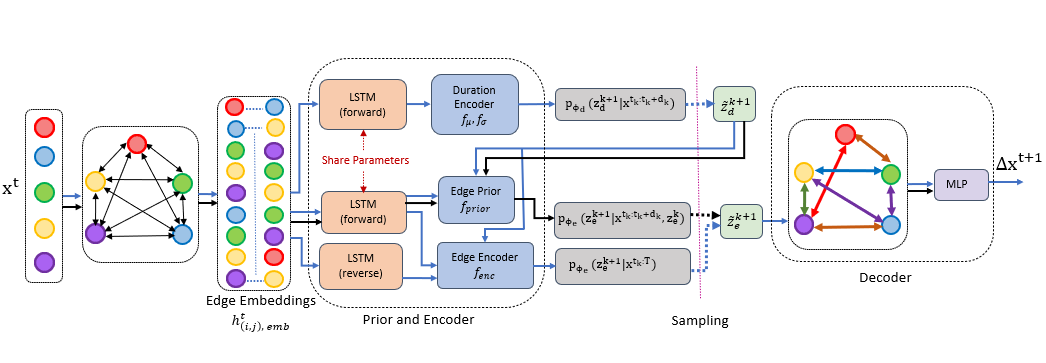}    \vspace{-0.75cm}
    \caption{{\textbf{Architecture of DIDER:} The input trajectories are fed to a fully-connected GNN to produce edge embeddings at every time step. These are aggregated using forward and reverse LSTM to encode the past and future trajectories. The duration prior and the edge prior are computed as a function of the past trajectory, and the edge encoding is computed as a function of both past and future. The edge types are sampled from the edge encoder during training and edge prior during inference. The decoder predicts the state of the entities at the next time step. The path shown by solid and dotted blue and black arrows corresponds to training and evaluation. }} 
    \vspace{-1.5em}
    \label{fig:arch}
\end{figure*}

\section{MODEL DESIGN}
Our objective is to discover an interpretable sequence of sub-interactions among agents from their observations in an online setting. We achieve this by disentangling the task of predicting dynamic interactions into two parts -sub-interaction prediction and duration estimation, both of which are unobserved. Thus, we model this problem using the Conditional Variational Autoencoders (CVAE) framework with two latent variables. The discrete and continuous latent variables $z_e$ and $z_d$ represent agents' interaction (edge) type and the corresponding time duration. Our model learns an unknown number of varying lengths of sub-interactions from the observations. The formulation involves the simultaneous prediction of time duration and interaction type between agents. This requires a novel encoder and prior models, as motivated by previous work on sequential segmentation modeling \cite{graber2020dynamic, tanneberg2021skid, gregor2018temporal}. The model is trained by maximizing ELBO. The architecture of DIDER is shown in Fig. \ref{fig:arch}.

In Dynamic NRI, the prior and encoder predict interaction at every time step $t$, by capturing past and past + future instances of trajectories, respectively. In contrast, DIDER firstly predicts the time duration of an interaction type with \textit{ Duration encoder} using the last segment of past trajectories. The time duration sampled from the duration prior is then used by Edge prior and Edge encoder to learn an interaction type corresponding to the specific segment of the trajectory. 

\vspace{-0.5em}
\subsection{Prior and Encoder}
\vspace{-0.5em}
To model evolving sequence of sub-interactions, we learn prior probabilities on the edge duration $z_d$ and edge types $z_e$ conditioned on the past. Similar to NRI and DNRI, the input at each time step is passed through a Graph neural network to produce edge embeddings, as follows:
\begin{equation}
    \begin{aligned}
h^t_{i, 1} = f_{emb}(x^t_i),
    \end{aligned}
\end{equation}
\begin{equation}
    \begin{aligned}
v \rightarrow e : h^t_{(i, j), 1} = f^1_e([h^t_{i, 1}, h^t_{j, 1}]),
    \end{aligned}
\end{equation}
\begin{equation}
    \begin{aligned}
e \rightarrow v : h^t_{(i, j), 2} = f^1_v(\Sigma_{i \neq j}h^t_{(i, j), 1}),
    \end{aligned}
\end{equation}
\begin{equation}
    \begin{aligned}
v \rightarrow e : h^t_{(i, j), emb}  =f^2_{emb}([h^t_{i, 2}, h^t_{j, 2}]).
    \end{aligned}
\end{equation}

This architecture implements a form of neural message passing in a graph where vertices $v$ represents entities $i$, and edges $e$ represents the relations between entities pairs $(i, j)$. $f_{emb}$, $f_e^1$ and $f_v^1$ are MLPs. The embeddings $h^t_{(i, j), 1}$ only depends on $x_i$ and $x_j$, while $h^t_{(i, j), 2}$ uses information from the whole graph. We refer to \cite{kipf2018neural, graber2020dynamic} for details. This neural message passing architecture outputs a per time step edge embedding $h^t_{(i, j), emb}$, which is fed into the forward and reverse LSTM networks to model the probabilities over edge duration and edge types. 
\begin{equation}
h^t_{(i, j), prior} = LSTM_{forward}(h^t_{(i, j), emb}, h^{t-1}_{(i, j), prior}),
\end{equation}
\begin{equation}
h^t_{(i, j), reverse} = LSTM_{reverse}(h^t_{(i, j), emb}, h^{t+1}_{(i, j), reverse}).
\end{equation}

\subsubsection{Duration Encoder}
The edge duration $z^k_d$ is modeled as a continuous latent variable, and determines the duration ($d_k$) of an interaction type for $k^{th}$ segment of an edge. With the initial burn-in period (observation period with groundtruth trajectory) of $T_{obs}$, it models the probability distribution of the duration ($d_1$) of first segment ($k=1$) of an interaction as $p_{\phi_d}(z_d^{1}|x^{1:T_{obs}})$, where $t_1 = 0$, $d_{0} = T_{obs}$, and $t_k$ and $t_k + d_k$ represent the start time and the end time of $k^{th}$ segment. $T_{remaining}$ is the duration of the remainder of the trajectory, which has not been utilized for duration estimation of previous sub-interactions. It is represented as $T- t_k - d_k$.

\begin{comment}
\begin{equation}
\begin{aligned}
p_\phi_d(z_d|x) := \prod_{k=1}^{K} p_\phi_d(z_d^{k+1}|x^{t_{k}:t_{k} + d_{k}}, z_d^{k}),
\end{aligned}
\label{eqn:duration_prior}
\end{equation}
\end{comment}

\paragraph{Parametization of Continuous latent Variables}
We parametrize $p_{\phi_d}(z^k_d|x)$ by a Gaussian distribution, i.e $p_{\phi_d}(z^k_d|x) = \mathcal{N}(\mu_k, \sigma^2_k)$, where $\mu_k$, and $\sigma_k^2$ are parameterized by neural networks. Let the prior be a Gaussian distribution with $p(z_d) = \mathcal{N}(\mu_0, \sigma_0^2)$. We use the reparametrization trick for the Gaussian distributed $z_d$ to sample the time duration factor $z^k_d = \mu_k + \sigma_k \epsilon$, where $\epsilon$ is an auxiliary noise variable $\epsilon  \sim \mathcal{N}(0, 1)$. Then the time duration of the $k^{th}$ segment is estimated as:
% \vspace{-2em}
\begin{equation}
\begin{split}
\mu_{k+1} = \tanh(f_\mu(h^{t_k + d_k}_{(i, j), prior})), \\
\sigma_{k+1} = \text{sigmoid}(f_\sigma(h^{t_k + d_k}_{(i, j), prior})), \\
p_{\phi_d}(z_d^{k+1}|x^{t_k:t_k + d_k}) = \mathcal{N}(\mu_{k+1}, \sigma_{k+1}^2), \\
z_d^{k+1} = \mu_{k+1} + \sigma_{k+1} \epsilon,\\
d_{k+1} = z_d^{k+1}  \cdot T_{remaining},
\end{split}
\end{equation}
% \begin{comment}
% \begin{equation}
%     \begin{aligned}
% \mu_{k+1} = \tanh(f_\mu(h^{t_k + d_k}_{(i, j), prior})), 
%     \end{aligned}
% \end{equation}
% \begin{equation}
%     \begin{aligned}
% \sigma_{k+1} = \text{sigmoid}(f_\sigma(h^{t_k + d_k}_{(i, j), prior})),
%     \end{aligned}
% \end{equation}
% \begin{equation}
%     \begin{aligned}
% p_{\phi_d}(z_d^{k+1}|x^{t_k:t_k + d_k}) = \mathcal{N}(\mu_{k+1}, \sigma_{k+1}^2),
%     \end{aligned}
% \end{equation}
% \begin{equation}
%     \begin{aligned}
% z_d^{k+1} = \mu_{k+1} + \sigma_{k+1} \epsilon,
%     \end{aligned}
% \end{equation}
% \begin{equation}
%     \begin{aligned}
% d_{k+1} = z_d^{k+1}  \cdot T_{remaining},
%   \end{aligned}
% \end{equation}
% \end{comment}
% \vspace{-0.2em}
where $f_\mu$, $f_\sigma$ are realized using MLPs.
% \begin{comment}
% \begin{equation}
% \mu_{k+1} = tanh(f_\mu(h^{T}_{(i, j), prior})), 
% \end{equation}
% \begin{equation}
% \sigma_{k+1} = sigmoid(f_\sigma(h^{T}_{(i, j), prior})),
% \end{equation}
% \begin{equation}
% p_{\phi_d}(z_d^{k+1}|x^{t_k:T}) = \mathcal{N}(\mu_{k+1}, \sigma_{k+1}^2)
% \end{equation}
% \begin{equation}
% z_d^{k+1} = \mu_{k+1} + \sigma_{k+1} \epsilon
% \end{equation}
% \begin{equation}
% d_{k+1} = z_d^{k+1}  \cdot T_{remaining}
% \end{equation}
% where $f_\mu$, $f_\sigma$ are realized using MLPs.
% \end{comment}

\subsubsection{Edge Prior and Edge Encoder}
The edge prior probabilities over edge types are modeled in an autoregressive manner. For each segment duration $d_{k+1}$ sampled from the duration encoder, the prior probabilities over edge types are conditioned on the relation type predicted in the previous segment ($z_e^{k}$)  as well as the sequence of observations in that segment, as following-
\begin{equation}
p_{\phi_e}(z^{k+1}_{(i, j)}|x^{t_k:t_k+d_k}, z_e^{k}) := \text{softmax}(f_{prior}(h^{t_k + d_k}_{(i, j), prior})),
\end{equation}
\begin{equation}
    \begin{aligned}
p_{\phi_e}(z_e|x) := \prod_{k=1}^K p_{\phi_e}(z_e^{k+1}|x^{t_k:t_k+d_k}, z_e^{k}).
    \end{aligned}
    \label{eqn:edge_prior}
\end{equation}
We encode the dependence of previous $z_{e}^k$ to next $z_e^{k+1}$ in the hidden state $h^{t_k+d_k}_{(i, j), prior}$. 

During training, the encoder computes the approximate distribution of edge types for every segment by using the information of the whole sequence (past segment and future). The true posterior over the latent space is a function of the future states of the observed variable \cite{fraccaro2016sequential}. Therefore, similar to DNRI, we use a reverse LSTM to capture future states of the sequence. The relational embedding $h_{(i, j), emb}^t$ is passed through a reverse LSTM and then concatenated with the results of forward LSTM to estimate posterior as follows:
\begin{equation}
q_{\phi_e}(z^{k+1}_{(i, j)}|x) := \text{softmax}(f_{enc}([h^{t_k + d_k}_{(i, j), reverse}, h^{t_k + d_k}_{(i, j), prior}]).
\end{equation}
The encoder approximates distribution of interactions for each segment as follows:
\begin{equation}
    \begin{aligned}
q_{\phi_e}(z_e|x) := \prod_{k=1}^K p_{\phi_e}(z_e^{k+1}|x^{t_k:T}).
    \end{aligned}
        \label{eqn:edge_encoder}
\end{equation}
The encoder and prior models share the parameters, so we use $\phi_e$ to refer to the parameters of both of these models.

\paragraph{Parameterization of Discrete latent Variables}
We parameterize discrete categorical distribution with a continuous approximation function (\textit{i.e.}, softmax) to obtain probability distribution over each edge type \cite{maddison2016concrete, jang2016categorical}. The sampling is done via reparametrization by first sampling a vector $g$ of independent and identically distributed samples drawn from Gumbel (0, 1) and computing the following,
\begin{equation}
\begin{aligned}
z_{e_{(i, j)}} = \text{softmax}(h{(i, j) + g}/ \tau),
\end{aligned}
\end{equation}
where $\tau$ is the softmax temperature which controls the sample smoothness. 

\subsubsection{Generic framework of Edge prior and Edge encoder}
We investigate the general formulation of edge prior and encoder modules, discussed in Eqn. \ref{eqn:edge_prior} and \ref{eqn:edge_encoder}. For a particular case where the number of segments $K$ is hardcoded as equal to time horizon $T$, corresponding to $d_k = 1$, the formulation of edge prior and encoder is factorized as:
\begin{equation}
    \begin{aligned}
p_{\phi_e}(z_e|x) := \prod_{t=1}^T p_{\phi_e}(z_e^{t+1}|x^{1:t}, z_e^{1:t-1}),
    \end{aligned}
    \label{eqn:generic_prior}
\end{equation}
\begin{equation}
    \begin{aligned}
q_{\phi_e}(z_e|x) := \prod_{t=1}^T p_{\phi_e}(z_e^{t+1}|x).
    \end{aligned}
        \label{eqn:genertic_encoder}
\end{equation}
This specific case corresponds to the encoder and prior formulation of DNRI. Therefore, DIDER provides a generic framework with a time duration encoder, which provides additional flexibility for improving the interpretability of models which follows a VAE-like framework, such as DNRI.

\subsection{Decoder}
Decoder is used to predict the trajectory given the observations of the entities, and the sampled relation types at every time step $t$. Similar to NRI and DNRI, we use an autoregressive model, which factorizes as follows:
\begin{equation}
p_{\theta}(x|z_e) := \prod_{t=1}^T p_\theta(x^{t+1}|x^{1:t}, z_e^{1:t}).
\end{equation}

We always provide the ground truth states to the decoder during training for the same reason suggested in  DNRI.

\subsection{Training and Inference}
We jointly train the generative and inference model parameters $\theta$, $\phi_d$ and $\phi_e$, by maximizing the ELBO:

\vspace{-0.5em}

\begin{equation}
\begin{split}
\mathcal{L}(\theta, \phi_d, \phi_e) = \mathbb{E}_{q_\phi}(z_e|x)[log p_{\theta_d}(x|z_e)] \\                - \beta_d (KL(q_{\phi_d}(z_d|x)||p(z_d)) - C_d) \\
- \beta_e (KL(q_{\phi_e}(z_e|x)||p_{\phi_e}(z_e|x)) - C_e).
\end{split}
\end{equation}
where $\beta_d$, $\beta_e$ are constant scaling factors, and $C_d$, $C_e$ are the information capacity terms. The first term aims at reconstructing the data, while the KL divergence forces the model to stay close to the given prior. Further, to enforce disentanglement, we use $\beta$-VAE formulation, as introduced in \cite{higgins2016beta}. Similar to SKID, we add capacity terms to the ELBO as proposed in \cite{dupont2018learning, burgess2018understanding}. Since DIDER discovers a sequence of sub-interactions for each edge individually, it adds a complexity of $O(n^2T)$, where $n$ is the number of agents.

%This balancing was further refined by adding capacity terms to the KL \cite{}, which can be seen as a slack variable allowing some distance to the KL, and which is increased during training. 

\begin{figure}
    \vspace{-1em}
    \centering
    \includegraphics[width=0.8\linewidth]{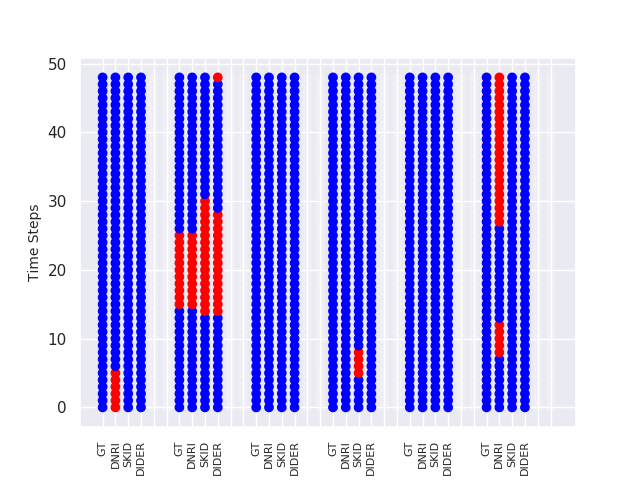}
    \caption{{Visualization of latent interactions inferred by DNRI, DIDER (with SKID), and DIDER (Ours) on synthetic physics simulations data with 3 particles. There exists 6 interactions for 3 particles' environment. Blue edge represents \textit{Non-Interacting} edge type, and red represents \textit{Interacting} edge type. GT denotes GroundTruth edges.}} 
    \label{fig:DNRI_DIDER_synth} 
    \vspace{-0.5em}
\end{figure}

\begin{figure}
    \vspace{-0.5em}
    \centering
    \includegraphics[width=0.8\linewidth]{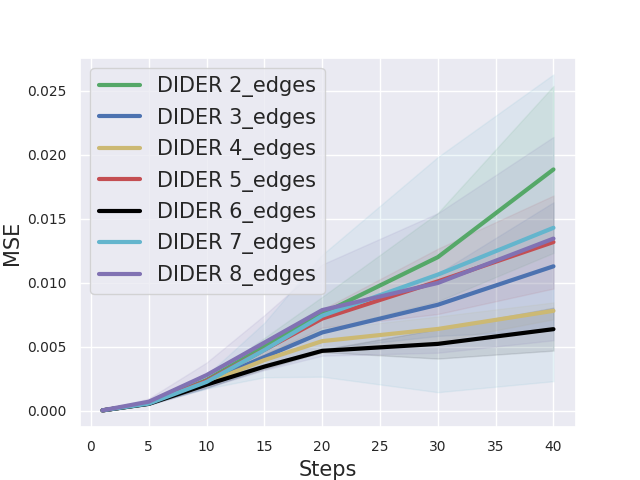}
    \caption{{Ablation study on DIDER with inD data, to determine the optimal number of edge types based on MSE loss of downstream trajectory prediction task}} 
    \label{fig:edges_ablation_dider}
    \vspace{-2em}
\end{figure}
%%%%%%%%%%%%%%%%%%%%%%%%%%%%%%%%%%%%%%%%%%%%%%%%%%%%%%%%%%%%%%%%%%%%%%%%%%%%%%%%

\begin{table*}[t]
  \caption{MSE and Edge Accuracy of Synthetic Data with 3 Particles}
 \begin{center}
  \begin{tabular}{|c|c|c|c|c|c|}
    \hline
    \multirow{2}{*}{Models} &
      \multicolumn{3}{c|}{MSE for various prediction horizons} &
      \multicolumn{2}{c|}{Edge accuracy (in $\%$)}\\
    & {1} & {15} & {25} & {\textit{No Interaction} edge} & {\textit{Interaction} edge} \\
    \hline
    DNRI & $(8.16 \pm 1.43) \times 10^{-7}$ &  $(3.19 \pm 1.79) \times 10^{-3}$ & $(4.18 \pm 2.25) \times 10^{-3}$ & $85.76 \pm 0.40$ & $58 \pm 9.80$\\
    
    \hline
    
    DIDER (with SKID) &  $(7.62 \pm 2.45) \times 10^{-7}$ &  $(1.63 \pm 0.38) \times 10^{-3}$ & $(2.20 \pm 0.50) \times 10^{-3}$ & $87.90 \pm 1.10$ & $\mathbf{94.08 \pm 0.50}$\\
    
    \hline
    
    DIDER (Ours) & $\mathbf{(4.71 \pm 0.80) \times 10^{-7}}$ &  $\mathbf{(1.13 \pm 0.17) \times 10^{-3}}$ & $\mathbf{(1.51 \pm 0.17) \times 10^{-3}}$ & $\mathbf{87.84 \pm 0.60}$ & $92.50 \pm 1.08$\\
    
    \hline
  \end{tabular}
  \end{center}
  \label{tab:synth}
\end{table*}

%%%%%%%%%%%%%%%%%%%%%%%%%%%%%%%%%%%%%%%%%%%%%%%%%%%%%%%%%%%%%%%%%%%%%%%%%%%%%%%%%%%%%%%%
\begin{table*}
  \caption{ MSE of Trajectory Prediction on Basketball Dataset with 5 agents}
  \centering
  \begin{tabular}{|l|c|c|c|}
    \hline
    \multirow{2}{*}{Models} &
      \multicolumn{3}{c|}{MSE for various prediction horizons}\\
     
    & {1} & {5} & {9}\\
    \hline
    DNRI & $(7.90 \pm 0.58) \times 10^{-5}$ &  $(5.98 \pm 0.12) \times 10^{-4}$ & $(2.42 \pm 0.10) \times 10^{-3}$\\
    
    \hline
    
    DIDER (with SKID) & $\mathbf{(7.35\pm 0.07) \times 10^{-5}}$ &  $(5.79\pm 0.20) \times 10^{-4}$ & $(2.38 \pm 0.06) \times 10^{-3}$\\
    
    \hline
    
    DIDER (Ours) & $(7.44 \pm 0.12) \times 10^{-5}$ &  $\mathbf{( 5.65\pm 0.08) \times 10^{-4}}$ & $\mathbf{(2.34\pm 0.03) \times 10^{-3}}$\\
    
    \hline
  \end{tabular}
  \label{tab:basket}
\end{table*}

%%%%%%%%%%%%%%%%%%%%%%%%%%%%%%%%%%%%%%%%%%%%%%%%%%%%%%%%%%%%%%%%%%%%%%%%%%%%%%%%%%%%%%%%%%%
\begin{table*}
  \caption{MSE and Consistency Analysis of inD dataset evaluated on 50 timestep}
  \centering
  \begin{tabular}{|l|c|c|c|c|c|c|c|c|}
    \hline
    \multirow{2}{*}{Models} &
      \multicolumn{3}{c|}{MSE for various prediction horizons} &
      \multicolumn{2}{c|}{Consistency Check (max $\%$ samples/edge type)}\\
     
    & {1} & {20} & {40} & Parked Vehicles & Non-Intersecting trajectories \\
    \hline
    DNRI & $(3.40 \pm 0.43) \times 10^{-5}$ &  $(8.5 \pm 1.60) \times 10^{-3}$ & $(32.20 \pm 8.40) \times 10^{-3}$ & $ 63.10 \pm 2.20$/1 & $46.31 \pm 1.53$/1\\
    
    \hline
    
    DIDER (with SKID) &  $\mathbf{(1.54 \pm 0.06) \times 10^{-5}}$ &  $(4.50 \pm 0.20) \times 10^{-3}$ & $(5.35 \pm 1.25) \times 10^{-3}$ & $96.22 \pm 0.52$/4 &  $63.24\pm 0.23$/1\\
    
    \hline
    
    DIDER (Ours) & $(1.63 \pm 0.08) \times 10^{-5}$ &  $\mathbf{(4.49 \pm 0.28) \times 10^{-3}}$ & $\mathbf{(5.27 \pm 1.64) \times 10^{-3}}$ & $\mathbf{99.89 \pm 0.41}$/3 &  $\mathbf{64.86\pm 0.24}$/1\\
    
    \hline
  \end{tabular}
  \label{tab:ind}
\end{table*}

%%%%%%%%%%%%%%%%%%%%%%%%%%%%%%%%%%%%%%%%%%%%%%%%%%%%%%%%%%%%%%%%%%%%%%%%%%%%%%%%%%%%%%%%%%%

\begin{table*}
  \caption{MSE of inD dataset evaluated for 200 timesteps}
  \centering
  \begin{tabular}{|l|c|c|c|c|c}
    \hline
    \multirow{2}{*}{Models} &
      \multicolumn{4}{c|}{MSE for various prediction horizons}\\
     
    & {1} & {40} & {120} & {190} \\
    \hline
    DNRI & $(5.67 \pm 0.42) \times 10^{-5}$ &  $(7.70 \pm 2.90) \times 10^{-2}$ & $2.37 \pm 1.52 $ &  $17.87 \pm 11.76$\\
    
    \hline
    
    DIDER (with SKID) &  $(3.28 \pm 1.49) \times 10^{-5}$ &  $(2.49 \pm 0.70) \times 10^{-2}$ & $ \mathbf{0.34 \pm 0.26}$ &  $3.97 \pm 5.58$\\
    
    \hline
    
    DIDER (Ours) & $\mathbf{(2.37 \pm 0.11) \times 10^{-5}}$ &  $\mathbf{(2.38 \pm 0.50) \times 10^{-2}}$ & $0.35 \pm 0.33 $ &  $\mathbf{1.96 \pm 2.18}$\\
    \hline
  \end{tabular}
  \label{tab:ind_500}
\end{table*}

\section{Experiments}
We demonstrate the performance of DIDER on the synthetic dataset, basketball dataset, and inD dataset \cite{bock2020ind}. The synthetic data was generated similar to as generated in DNRI \cite{graber2020dynamic}. We mainly compare the performance with DNRI and SKID, which are mostly related to us. Originally, SKID is not designed for segmenting interactions from multiagent observations, but we are adopting their framework to learn to segment interactions from multiagent interactions. Further, SKID uses complete trajectory information for predicting the segment duration, making it suitable for offline setting. As we consider trajectory prediction as a surrogate task, we aim to perform disentangled interaction segmentation in an online setting. Therefore, we do not assume access to future observations during evaluation. We include an ablation study, where we use SKID-based \textit{Duration Encoder} for predicting the interaction duration in DIDER by providing complete trajectory information. We use the following labels for our proposed framework, baseline method, and ablation study- 
\begin{itemize}
    \item DNRI: Dynamic Neural Relational Inference  \cite{graber2020dynamic}
    
    \item DIDER (Ours): Our proposed framework  
    
    \item DIDER (with SKID): Our framework, with \textit{Duration Encoder} formulation adapted from SKID \cite{tanneberg2021skid} 
    
\end{itemize}

For all experiments, the first edge type is hardcoded to represent \textit{No-Interaction}. We conduct four statistically independent runs with random seeds from 1 to 4 and report the mean and standard deviation. The plots present the average performance with the shaded region showing a $95\%$ confidence interval. To demonstrate the interpretability of discovered interactions, we either compute edge accuracy, visualize the interaction transitions across time, or both.

\subsection{Synthetic Physics Simulation}
To evaluate the performance of DIDER and showcase its ability to discover dynamic relations, we use physical simulation systems, \textit{i.e.}, moving particles with dynamic relations between them \cite{graber2020dynamic}. As the dataset is synthetically generated, the ground truth interactions are known in prior as: \textit{Interaction} and \textit{No-Interaction}. The dataset consists of three particles: where two particles move with constant velocity, and the third is initialized with a random velocity. The third one is pushed away by other particles whenever the distance separating them is less than 1, and its edge type changes from \textit{No-Interaction} to \textit{Interaction}. We generated 40k samples with the trajectory length of 50 steps.

During the evaluation, we provide the ground truth position and velocity corresponding to the first 5 steps, and the models predict the remainder of the trajectory. 
 Since we have access to the ground truth interaction type for this dataset, we compare the edge accuracy corresponding to both the edge types. Our findings are summarized in Table \ref{tab:synth}. Results demonstrate that DIDER (Ours) outperforms DNRI and DIDER (with SKID) in MSE and accuracy. We further show that DIDER (Ours) shows comparable performance with DIDER (with SKID) for \textit{Interaction} edge type accuracy.  
 
 %We do not observe any significant difference in the performance between DIDER (Ours) and with SKID based Duration encoder, mainly because the trajectories duration is shorte
 % Further, we evaluate the performance of DIDER with SKID type duration estimation module, and results demonstrate that DIDER outperforms DIDER (with SKID) on edge accuracy. 
 
We also provide the visualization of 6 edges corresponding to 3 particle simulations data predicted by all methods in Fig \ref{fig:DNRI_DIDER_synth}. DIDER (with SKID) and DIDER (Ours) are very close to ground truth edge types compared to DNRI. It also shows that the DIDER can predict interactions having only one edge type as \textit{Non-Interacting} type across time.

%\begin{table*}[t]
%  \caption{\label{tab:synth} MSE and Edge Accuracy of Synth Dataset with 3 Particles and 5 Particles}
%  \centering
%  \begin{tabular}{l|c|c|c|c|c|c|c|l}
%    \hline
%    \multirow{2}{*}{Methods} &
%      \multicolumn{4}{c}{3 Particles} &
%      \multicolumn{4}{c}{5 Particles}\\
%    & {1} & {15} & {25} & {Edge Accuracy} & {1} & {15} & {25} & {Edge Accuracy} \\
%    \hline
%    DNRI & $6.6 \times 10^{-7}$ & 0.0012 & 0.0015 & 86/70 & $3.34 \times 10^{-6}$ & 0.0017 & 0.0024 & 89/43\\
%    \hline
%    DIDER (with SKID) & \textbf{$2.94 \times 10^{-7}$} & 0.0008 & 0.0012 & 89/91 &  \\
%    \hline
%    DIDER & $3.5 \times 10^{-7}$ & \textbf{0.00024} & \textbf{0.0003} & 91/84& $1.48 \times 10^{-6}$& 0.00053 & 0.00086 & 92/25\\
%    \hline
%  \end{tabular}
%\end{table*}

%%%%%%%%%%%%%%%%%%%%%%%%%%%%%%%%%%%%%%%%%%%%%%%%%%%%%%%%%%%%%%%%%%%%%%%%%%%%%%%%%%

\begin{comment}
\begin{figure}
    \centering
    \includegraphics[width=\linewidth]{images/DNRI_DIDER.png}
\caption{{Trajectory prediction errors on inD dataset}} 
    \label{fig:DIDER_DNRI_ind}
\end{figure}
\end{comment}

\subsection{BasketBall Dataset}
The Basketball dataset \cite{yue2014learning} consists of trajectories (positions and velocities) of 5 players. The data is preprocessed into 49 frames which span approximately 8 seconds of play \cite{graber2020dynamic}. We train the model with initial 40 frames and are tasked to predict the remaining trajectories. The model uses two edge types as \textit{No-Interacting} and \textit{Interacting}. The results are summarized in Table \ref{tab:basket}. DIDER (Ours) and DIDER (with SKID) outperform DNRI. It demonstrates the efficacy of our proposed disentangled framework against the step-wise counterpart. Further, DIDER (Ours) outperforms DIDER (with SKID) for longer horizon predictions, i.e., for 5 steps and 9 steps. We do not have ground truth interactions and are unaware of human-defined semantics of interaction types for the basketball dataset, unlike the road user traffic dataset, where the semantics of interactions are typically defined as yielding, following, passing-by, etc. \cite{zhang2020multi}. Therefore, we only compare the performance using MSE.

\subsection{inD Dataset}
%inD (Intersection Drone dataset) is a naturalistic German intersections-based road user trajectory dataset. The dataset consists of four intersections and includes trajectories of more than 11500 road users, consisting of heterogeneous agents, such as cars, trucks, buses, pedestrians, and bicyclists. This dataset is suitable for our problem setting, as it incorporates various temporal dynamic interactions between agents, such as \textit{car-following}, \textit{yielding}, \textit{passing-by}, \textit{cutting-in}, \textit{lane-changing} etc. However, annotating dynamic interactions at the frame level is a non-trivial task and can be challenging and time-consuming. Further, it may introduce the annotator's subjectivity as well as bias. We aim to discover these latent interpretable dynamic interactions from agents' observations.

inD (Intersection Drone dataset) is a naturalistic German intersections-based road user trajectory dataset. This dataset is suitable for our problem setting, as it incorporates various temporal dynamic interactions between agents, such as \textit{car-following}, \textit{yielding}, \textit{passing-by}, \textit{cutting-in}, \textit{lane-changing} etc. However, per-step annotation of dynamic interactions between agents is a challenging and time-consuming task and may introduce annotator's subjectivity and bias. In this experiment, we aim to discover the latent interpretable sequence of dynamic interactions from agents' observations, also referred to as traffic-primitives in \cite{zhang2020multi}.

\begin{figure*}[hbt!]
    \centering
   \subfloat[\centering Noisy per-step interaction prediction by  DNRI]{{\includegraphics[width=0.40\linewidth]{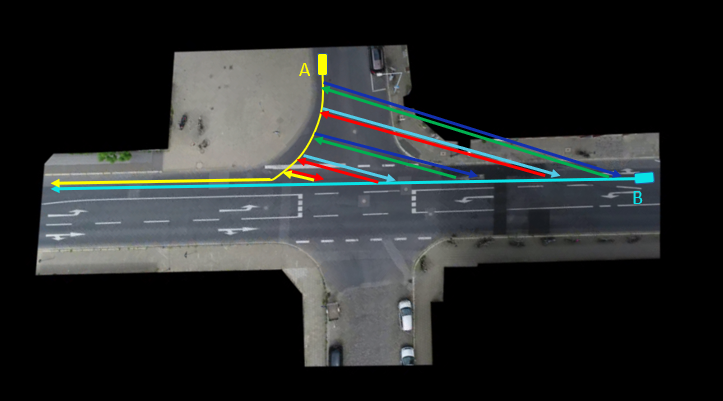} }\label{fig:yielding_dnri}}
    \qquad
    \subfloat[\centering Segments of sub-interactions prediction by DIDER ]{{\includegraphics[width=0.40 \linewidth]{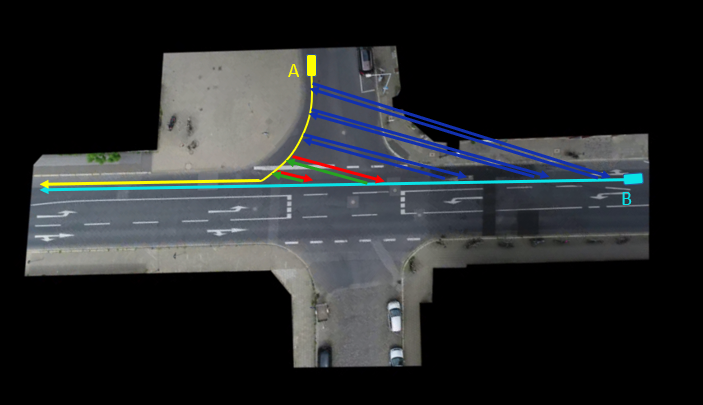} \label{fig:yielding_dider}}}
    \qquad
    \subfloat[\centering Visualization of yielding trajectories with DNRI]{{\includegraphics[width=0.40\linewidth]{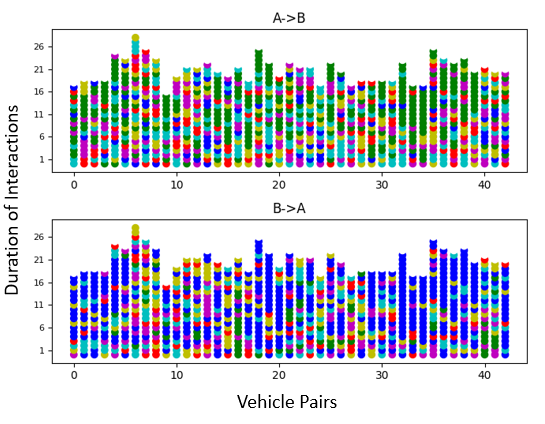} }\label{fig:yielding_dnri_dist}}
    \qquad
    \subfloat[\centering Visualization of yielding trajectories with DIDER ]{{\includegraphics[width=0.40\linewidth]{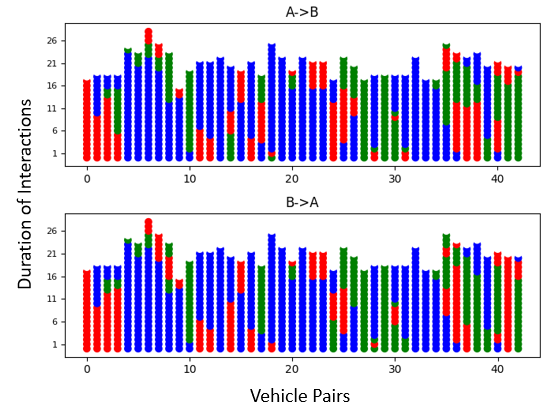} \label{fig:yielding_dider_dist}}}
   
     \caption{Qualitative Results on inD dataset: (a, b) Evolution of dynamic interactions between 2 cars approaching an intersection, where yellow car (A) yields for the straight going cyan car (B). The segments of interactions from $A \rightarrow B$ and $B \rightarrow A$ discovered by DIDER is shown by different colors. Based on these segments of interactions, we may infer that red represents \textit{yielding} interaction, and green represents \textit{passing-by} interaction, while blue is already hardcoded as \textit{no-interaction}.
     (c, d) Interactions segments corresponding to different pairs of cars involved in similar intersection at different times.}
    \label{fig:dnri_dider_ind}
    \vspace{-2em}
\end{figure*}

The dataset consists of 33 recordings, and we use a split of 19, 7, and 7 for the train, validation, and test dataset, respectively, as used in DNRI \cite{graber2020dynamic}. During testing, we divide the trajectories into sequences of 50 steps. Contrary to the previous dataset we used in this paper; the inD dataset has a varying number of agents at every time step. Therefore, for each agent present in the sequence, we provide the model with its ground truth position and velocity for the first 5 time steps, and the model forecasts the remainder of the trajectory.

As we do not have ground truth interactions for this dataset, we cannot directly measure the accuracy of each edge type, to quantify the interpretability. Moreover, it is challenging to determine the number of edge types $(e)$ for this dataset. Therefore, we conduct an ablation study with DIDER using various edge types, from 2 to 8 and select the one with the least MSE for the downstream trajectory prediction task. Fig. \ref{fig:edges_ablation_dider} shows the results of the ablation study with different edge types, and we choose $e = 6$ for all methods. The comparison of prediction task results with different methods is shown in Table \ref{tab:ind}. We further evaluate the performance of these methods on trajectories with longer sequences to highlight the difference in the performance of DIDER (with SKID) and DIDER (Ours). As the SKID-based framework uses an entire future sequence of trajectories to estimate the time duration of the current edge type, it adds noise into the estimation, which results in an inaccurate prediction of segment duration as sequence length increases. To demonstrate this, we train the models on inD dataset with longer sequences, where we divide trajectories into sequences of 200 steps and provide the ground truth for the first 5 steps. Table \ref{tab:ind_500} shows that DIDER (Ours) outperforms DIDER (with SKID) for longer sequences. 

%However, this paper aims not to compare the performance of DIDER (Ours) with DIDER (with SKID) but to develop an unsupervised framework for discovering interpretable interactions from multiagent observations.  

We further investigate the interpretability of these models on different traffic scenarios extracted from the dataset. For a sanity check, we firstly perform consistency analysis on the predicted interaction types between non-stationary vehicles, which seem to have no interactions based on our understanding of the scenarios. We extract these pairs of vehicles based on their location at the intersection, if they are moving in opposite directions lanes and their trajectories do not intersect at any time. The results show that DNRI, DIDER (with SKID) and DIDER (Ours) cluster $46.31\%$, $63.24\%$ and $64.86\%$, respectively, of such interactions under edge type 1, which is hard-coded to represent \textit{No-interaction}. Similar analysis of interactions between parked cars in the dataset shows that DNRI, DIDER (with SKID), and DIDER (Ours) cluster a maximum number of such interactions under edge types 1, 4, and 3, respectively, with their percentage as $63.10\%$, $96.22\%$, and $99.89\%$, though we don't have the semantic meaning of these edge types.

We further visualize the interactions corresponding to pair of cars, where one vehicle yields for the other vehicle in the scene, as shown in Fig. \ref{fig:yielding_dnri}. The distribution of edge types shows that DIDER predicts a sequence of sub-interactions that are consistent across different pairs of cars, as shown in Fig. \ref{fig:yielding_dider_dist}, while DNRI predicts noisy interactions, as shown in Fig. \ref{fig:yielding_dnri_dist}. The difference in the sequence of sub-interactions across various pairs of cars is due to the different times they approach the intersection. 

%%%%%%%%%%%%%%%%%%%%%%%%%%%%%%%%%%%%%%%%%%%%%%%%%%%%%%%%%%%%%%%%%%%%%%%%%%%%%%%%
\section{CONCLUSION}
This paper introduces DIDER, a generic, intrinsically interpretable, and unsupervised framework that learns disentangled and interpretable inter-agent relations from unlabeled raw observations. It targets a class of tasks where agents interact with each other in a sequence of temporal sub-interactions. DIDER achieves this by using discrete and continuous latent space to disentangle dynamic interactions prediction into sub-interaction and duration prediction. We demonstrate that DIDER achieves interpretable relations, as well as better trajectory prediction performance, as compared to models that predict interactions at every time step. 

% An interesting future direction of this work is to investigate the problem with a more complex autonomous driving dataset with infrastructure, context, and heterogeneous agents. Further, we would investigate how the learned causal graphs can provide human interpretable explanations for predictions. 

An interesting future direction is to investigate the problem with a more complex autonomous driving dataset with infrastructure, context, and heterogeneous agents. Further, we would study how the learned causal graphs can provide human interpretable explanations for predictions.

%%%%%%%%%%%%%%%%%%%%%%%%%%%%%%%%%%%%%%%%%%%%%%%%%%%%%%%%%%%%%%%%%%%%%%%%%%%%%%%%

%%%%%%%%%%%%%%%%%%%%%%%%%%%%%%%%%%%%%%%%%%%%%%%%%%%%%%%%%%%%%%%%%%%%%%%%%%%%%%%%

%%%%%%%%%%%%%%%%%%%%%%%%%%%%%%%%%%%%%%%%%%%%%%%%%%%%%%%%%%%%%%%%%%%%%%%%%%%%%%%%
\bibliographystyle{IEEEtran}
\bibliography{reference}

\end{document}